\pdfoutput=1
\documentclass[11pt]{article}

\usepackage[]{ACL2023}

\usepackage{times}
\usepackage{latexsym}

\usepackage[T1]{fontenc}
\usepackage[utf8]{inputenc}

\usepackage{microtype}

\usepackage{inconsolata}

\usepackage{booktabs}
\usepackage{inconsolata}
\usepackage{times}
\usepackage{color}
\usepackage{latexsym}
\usepackage{amsmath}
\usepackage{booktabs}
\usepackage{multirow}
\usepackage{graphics}
\usepackage{graphicx} 
\usepackage{caption}
\usepackage{amssymb}
\usepackage{pifont}
\usepackage{subfigure}
\usepackage{bm}
\usepackage{bbding}
\usepackage{float}
\usepackage{amsmath}
\usepackage{colortbl}  %彩色表格需要加载的宏包
\usepackage{xcolor}
\usepackage{tcolorbox} %附录里prompt template需要的包
\usepackage{inconsolata}
\usepackage{soul}
\newcommand{\ctext}[3][RGB]{%
  \begingroup
  \definecolor{hlcolor}{#1}{#2}\sethlcolor{hlcolor}%
  \hl{#3}%
  \endgroup
}

\usepackage{makecell}
\usepackage{hyperref}
\newcommand{\model}{HAI}
\definecolor{DarkGreen}{RGB}{0, 128, 0}   % 深绿色
\begin{document}

% \title{{\model}: Automatic Agent Prompts Optimization for Tool-augmented Chart-based Visual Question Answering}
\title{Chart-HQA: A Benchmark for Hypothetical Question Answering in Charts}

% \author{Xiangnan~Chen, Yuancheng~Fang, Qian~Xiao, Juncheng~Li, Jun Lin, Siliang~Tang, \\ Yi Yang,~\IEEEmembership{Senior~Member,~IEEE}, Yueting~Zhuang,~\IEEEmembership{Senior~Member,~IEEE}
% \IEEEcompsocitemizethanks{\IEEEcompsocthanksitem X. Chen, Y. Fang, J. Li, S. Tang, Y. Yang, Y. Zhuang
%     are with the College of Computer Science and Technology, Zhejiang University, China. E-mail:\{ xnchen2020; ycfang; junchengli; siliang; yangyics; yzhuang\}@zju.edu.cn
% 		\IEEEcompsocthanksitem Q. Xiao, J, Lin are with the Alibaba Group, Hangzhou, China. E-mail:\{ xiaoqian.xq; linjun.lj\}@alibaba-inc.com}}

\author{
Xiangnan Chen$^{1}\thanks{~~Work done during an internship at DAMO Research, Alibaba Group.}$ , 
Yuancheng~Fang$^{1}$,
{\bf Qian Xiao}$^{2}$, Juncheng Li$^{1}$ 
 ,  {\bf Jun Lin}$^{2}$,\\ 
 \textbf{Siliang Tang}$^{1}$,
 \textbf{Yi Yang}$^{1}$,
 \textbf{Yueting Zhuang}$^{1}$,
 \\
 $^{1}$ Zhejiang University 
$^{2}$ Alibaba Group\\
  \texttt{\{xnchen2020,ycfang,junchengli,siliang,yangyics,yzhuang\}@zju.edu.cn}
  \\
  \texttt{\{xiaoqian.xq,linjun.lj\}@{alibaba-inc.com}
  } }

% \input{Sections/1_Introduction}
% \input{Sections/2_Related_Work}
% \input{Sections/3_Methodology}
% \input{Sections/9_Dataset}
% \input{Sections/4_Experiment}
% \input{Sections/10_Analysis}
% \input{Sections/5_Conclusion}
% % \input{Sections/6_Limitation}
% % \begin{thebibliography}{1}
% \bibliographystyle{IEEEtran}
% % \bibliographystyle{ACM-Reference-Format}
% \bibliography{citation}
% \clearpage
% \input{Sections/7_Appendix}

% 
\maketitle
\begin{abstract}
Multimodal Large Language Models (MLLMs) have garnered significant attention for their strong visual-semantic understanding. Most existing chart benchmarks evaluate MLLMs’ ability to parse information from charts to answer questions. However, they overlook the inherent output biases of MLLMs, where models rely on their parametric memory to answer questions rather than genuinely understanding the chart content. To address this limitation, we introduce a novel Chart Hypothetical Question Answering (HQA) task, which imposes assumptions on the same question to compel models to engage in counterfactual reasoning based on the chart content. Furthermore, we introduce {\model}, a human-AI interactive data synthesis approach that leverages the efficient text-editing capabilities of LLMs alongside human expert knowledge to generate diverse and high-quality HQA data at a low cost. Using {\model}, we construct Chart-HQA, a challenging benchmark synthesized from publicly available data sources. Evaluation results on 18 MLLMs of varying model sizes reveal that current models face significant generalization challenges and exhibit imbalanced reasoning performance on the HQA task. 
% Our codebase and newly generated datasets are available at \href{https://anonymous.4open.science/r/Chart-HQA-86BE}{\textcolor{blue}{https://anonymous.4open.science/r/Chart-HQA-86BE}}.
\end{abstract}

\section{Introduction}
\label{sec:intro}
% 1.说明多模态大模型在图表上的能力还有待研究
% 2。当前图表benchmark的特点和问题
% 3. 我们提出假设性问答，特点 优势是啥
% 4. 为了成本 想用机器生成HQA 有两个挑战
% 5. 为了解决挑战，提出我们的数据构造方法
% 6. 构造了Chart-HQA，有哪些特点 和实验发现
% 最后总结全文贡献

% With the breakthrough advancements in Large Language Models (LLMs)~\cite{clip,GPT-3,PaLM,llama}, 
% Multimodal Large Language Models (MLLMs)~\cite{BLIP2,llava,Minigptv1} 
% have emerged as a dominant approach for multimodal learning. 
% These models leverage powerful LLMs as core reasoning engines to perform multimodal tasks.

Multimodal Large Language Models (MLLMs)~\cite{BLIP2,llava} have demonstrated exceptional performance in visual-semantic understanding~\cite{GPT4,cogvlm}.
% such as zero-shot instruction following~\cite{li2023finetuning} and OCR-free mathematical reasoning~\cite{zhu2023minigpt4,dai2023instructblip}.
Despite their success, existing MLLMs still face significant challenges in reading, understanding, and summarizing visual charts~\cite{ChartQA,sciGraphQA}. Unlike natural images, which primarily rely on recognizable objects, relative positions, and interactive relationships to convey information, charts communicate complex semantic meanings through visual logic~\cite{chartbench}, such as trend lines, color-coded legends, and axis structures. 

% Since charts present detailed and highly structured data in a visual format, 
% it is crucial to assess MLLMs' ability to understand visual charts.

\begin{figure*}[t]
    \centering
    % \small
    \includegraphics[width=1\textwidth]{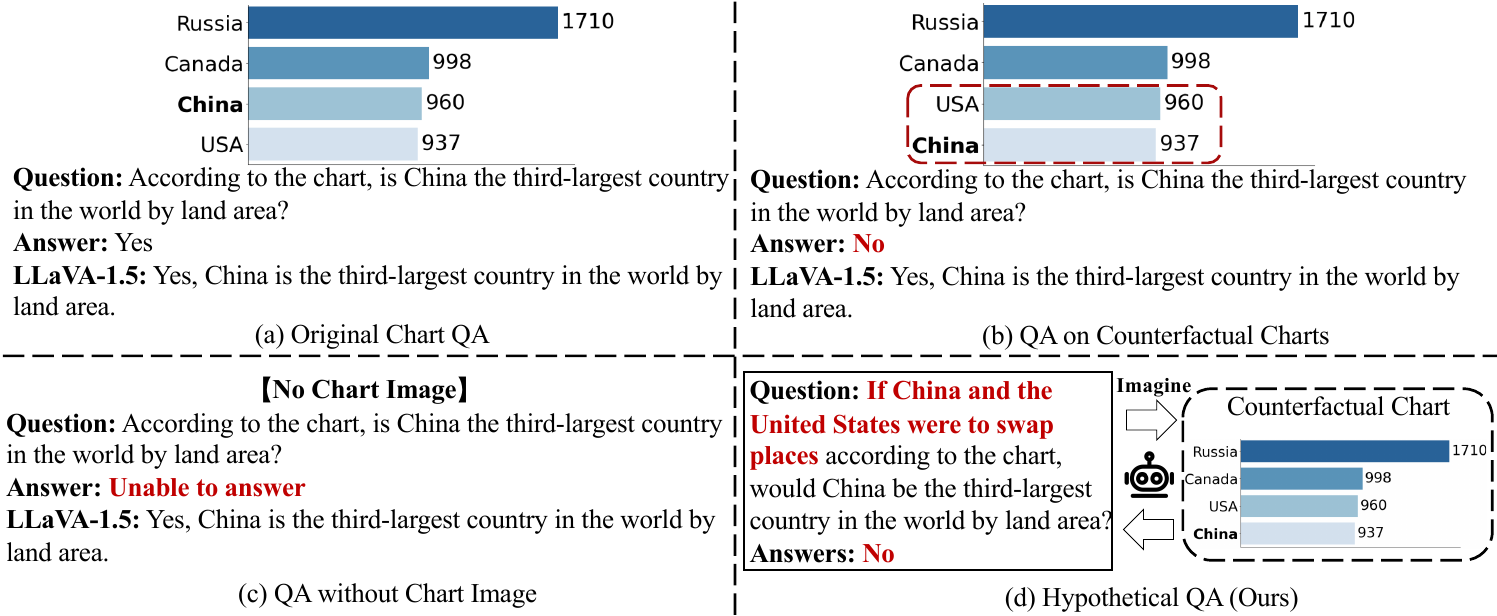} \\
    \caption{An example of biased output on charts from MLLMS and proposed hypothetical QA task.
    (a) Factoid  QA results based on the original chart.
    (b) The response after counterfactual editing of the chart, where the land areas of "China" and "USA" are swapped.
    (c) The model's answers without the chart image input.
    (d) Illustration of hypothetical question and the corresponding counterfactual context to be imagined.}
    \label{hqa:biased}
    \vspace{-4mm}
\end{figure*}

Most existing chart question answering benchmarks~\cite{ChartQA, ChartX} mainly focus on factoid question answering (FQA), 
where the model is required to directly extract information from the
chart image to answer question, as shown in Figure~\ref{hqa:biased}.a.
Although these benchmarks have made significant progress in expanding dataset scale~\cite{xia2024chartvlm} and diversifying chart types~\cite{chartbench}, \textbf{they overlook the inherent output biases problem of MLLMs}~\cite{huang-etal-2024-lvlms,HallusionBench}, i.e., MLLMs tend to rely on their parametric memory to answer questions rather than interpreting the visual content of the chart. Taking the widely used multimodal model 
LLaVA-1.5~\cite{llava1.5} as an example, as shown in Figure~\ref{hqa:biased}, although LLaVA-1.5 correctly answered the question in the FQA task (Figure~\ref{hqa:biased}.a), the model still produced the same output when the counterfactual image was provided (Figure~\ref{hqa:biased}.b) or even the chart image was missing (Figure~\ref{hqa:biased}.c).
This phenomenon indicates that introducing additional control conditions (e.g., missing images, counterfactual images) for the same chart question can effectively reveal the output bias of MLLMs, thereby reflecting their true understanding of charts. 
Unfortunately, to the best of our knowledge, no existing chart benchmarks have been designed to thoroughly investigate such problem.

% assess MLLMs' ability to understand visual charts.

% \textbf{Due to the output biases, existing chart FQA benchmarks struggle to accurately reflect the models’ true understanding of visual charts.} Taking the widely used model 
% LLaVA-1.5~\cite{llava1.5} as an example, as shown in Figure~\ref{hqa:biased}, although LLaVA-1.5 correctly answered the question in the FQA task (Figure~\ref{hqa:biased}.a), the model still produced the same output when the chart image was missing (Figure~\ref{hqa:biased}.c) or even the counterfactual image was provided (Figure~\ref{hqa:biased}.b).
% This phenomenon suggests that incorporating additional control conditions for the same chart question can effectively reveal the true understanding capability of MLLMs.
% das

% 据我所知，目前还没有benchmark去深入探讨认知偏差问题
% 为了填补这一空白，我们提出了一个全新任务，以反事实思考为角度去揭示模型在图表上的认知偏差。具体来说，任务介绍

To fill this gap, we propose an novel \textbf{hypothetical question answering} (HQA) task in the domain of chart understanding. Unlike directly modifying chart images as a control condition, we focus on imposing an assumption on the original chart question. As shown in Figure~\ref{hqa:biased}.d, the proposed HQA task requires models to independently \textit{imagine} the corresponding counterfactual details based on the given assumption and original chart image, thereby establishing an accurate inference context. 
The HQA task will undoubtedly enhance the practical use of MLLMs due to the universality of hypothetical questions in real-world scenarios.

% 一个朴素的方法人工设计 缺点1，
% 用机器缺点2同一假设不能适配不同的图表
% 两个缺点再精简些 详细内容放到方法章节
However, constructing a high-quality chart HQA benchmark is not trivial. While a straightforward approach is to utilize human experts for data synthesis, existing research~\cite{self-instruct} has shown that human-generated data suffer from \textbf{limited diversity}. Specifically, most human-generated hypothetical questions tend to focus on common chart attributes such as specific data, falling short of covering a true variety of assumption types and different ways to describe them. 
Secondly, the same hypothetical scenario may not be applicable to different chart types and could even lead to \textbf{conflicting layout structures}. For example, the assumption “\textit{suppose a specific value in the chart doubles}” is reasonable in a bar chart. However, in a pie chart, this assumption violates the structural constraint that all slices must sum to 100\%. Undoubtedly, such structurally conflicting hypothetical questions significantly reduce the practical applicability of the HQA benchmark.

To overcome above challenges, we propose a human-machine interactive HQA data synthesis method named \textbf{HAI}. HAI combines the efficient text editing capabilities of LLMs with human expert knowledge to synthesize diverse and high-quality HQA data at a low cost. Specifically, 
HAI consists of two key components: \textbf{(1) Counterfactual proposal generator} (CIG). To diversify counterfactual assumptions, the CIG module randomly samples a subset of instructions from the seed instruction set (initially composed of limited manual instructions) and inputs them into the LLM along with the detailed description of charts to generate new instruction proposals and HQA instances.
\textbf{(2) Human-feedback discriminator} (HFD). This module employs multiple human experts to review generated HQA instances from various perspectives, including answer accuracy, layout consistency, and question clarity. Subsequently, HQA instances validated by human experts are retained. Furthermore, leveraging the self-reflection capability~\cite{shinn2023reflexion} of LLMs, the corresponding instruction proposals are revised based on human expert feedback, thereby expanding the seed instruction set.
\begin{figure*}[t]
    \centering
    % \small
\includegraphics[width=\textwidth]{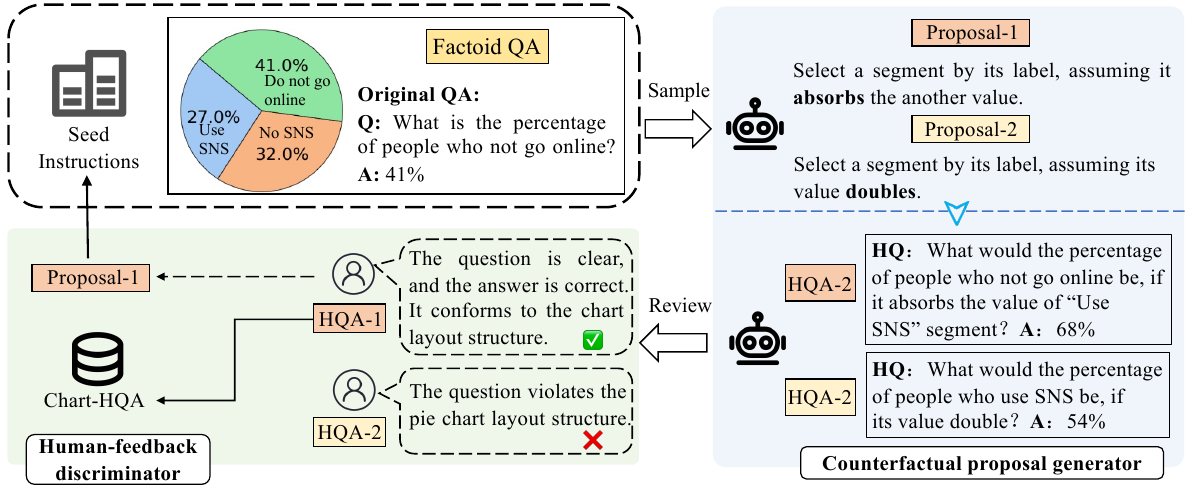}
    \caption{The illustration of our approach for synthesizing hypothetical questions, including two stages that synthesize new instruction proposals,  and human verification.}
    \label{fig:dataset_generation}
    \vspace{-2mm}
\end{figure*}
    
Based on the proposed method, we construct \textbf{Chart-HQA}, 
a challenging HQA benchmark derived from factoid QAs in ChartQA~\cite{ChartQA}. There are \textbf{\textit{900}} counterfactual instruction proposals and \textbf{\textit{4}} answer types within Chart-HQA. 
We evaluate the zero-shot reasoning capabilities of \textbf{18} MLLMs of varying model sizes on Chart-HQA, including \textbf{8} specialist chart-based models and \textbf{10} generalist models. The results (as shown in Table~\ref{tab:hqa}) indicate that existing models generally exhibit limited reasoning capabilities in chart hypothetical question answering. For instance, the high-performing GPT-4o~\cite{GPT4v} experiences a significant performance drop, with relaxed accuracy decreasing from \textit{85.7\%} on ChartQA to \textit{62.52\%} on Chart-HQA. Additionally, we observe that most models demonstrate imbalanced performance across different answer types within Chart-HQA, highlighting potential avenues for optimizing future MLLMs.
% We evaluate the zero-shot reasoning capabilities of \textbf{6} specialist chart-based visual question answering models and \textbf{5} generalist MLLMs on Chart-HQA. The results demonstrate that existing models generally exhibit limited reasoning capabilities when it comes to chart hypothetical question answering. For instance, the high-performing GPT-4V~\cite{GPT4v} experiences a significant performance drop, with relaxed accuracy falling from \textit{78.5\%} on ChartQA to \textit{56.49\%} on Chart-HQA. Additionally, we observe that most models demonstrate imbalanced performance across different answer types within Chart-HQA, highlighting potential avenues for optimizing future MLLMs. 

In summary, our main contributions are as follows:

\begin{itemize}
    \item We propose a novel chart hypothetical question answering task to evaluate the true understanding capabilities of MLLMs on chart-based reasoning.
    \item We propose a human-machine interactive HQA data synthesis framework named {\model}. It leverages LLMs to automatically generate diverse and high-quality HQA data under the guidance of human feedback.
    \item We unveil Chart-HQA, a challenging benchmark for chart hypothetical question answering. Extensive experiments demonstrate
    that existing MLLMs are generally inadequate in counterfactual chart comprehension abilities.
\end{itemize}
\section{Human-machine Interactive Data Synthesis}
In this section, we detail our unique method that synthesizes high-quality hypothetical QA for chart-based visual question answering, named {\model}. 
As shown in Figure~\ref{fig:dataset_generation},
our method consists of two interconnected modules, a Counterfactual proposal generator (CPG), and a Human-feedback discriminator (HFD). In section ~\ref{sec:proposal}, we present how the CPG module iteratively generate new instruction proposals and HQA instances.
In section~\ref{sec:human}, we introduce the HFD module to leverage human expert knowledge for validating HQA instances.
% Our method comprises three interconnected steps: proposal synthesis, hypothetical QA generation, and human verification. We illustrate this process in Figure~\ref{fig:dataset_generation}. These modules are detailed in the following subsections. All prompt templates are shown in Appendix~\ref{app:hqa_template}.
\subsection{Counterfactual Proposal Generator}
\label{sec:proposal}
% Our primary goal is to use machines to automatically generate diverse and high-quality HQA. However, using LLMs directly to annotate large-scale HQA data is not an easy task, because it requires (1) creativity to come up with novel counterfactual operations over the rich attributes of charts. (2) expertise in writing assumptions that are consistent with the logic of the question. Empirically, even if we give detailed task descriptions and provide some demonstrations, the model tends to repeat the reference data and lacks valuable insights that could be important for designing diverse HQA data. Therefore, in this step, we inspire LLMs to design the general instruction proposal, that describes a general counterfactual operation for charts. Inspired by the success of SELF-INSTRUCT~\cite{wang-etal-2023-self-instruct}, we generate a diverse set of instruction proposals in a bootstrapping fashion. First, we initialize an instruction proposal pool (4 seed proposals per chart type written by our authors). Secondly, given a set of general descriptions of chart attributes $\{D_C\}$, we employ GPT-4 to generate the counterfactual instruction proposals $I_P$, denoted as:
% \begin{equation}
% \begin{aligned}
% &I_P = \textit{GPT-4}(I_S, P_I, D_C), 
% \end{aligned}
% \end{equation}
% where $I_S$ is the seed instruction proposals sampled from the instruction proposal pool as in-context examples and $P_I$ is the prompt to guide GPT-4.
The primary goal of our method is to automatically generate diverse and high-quality Hypothetical Question Answering (HQA) instances using large language models (LLMs). However, directly leveraging LLMs to annotate large-scale HQA data is challenging, as it requires:
(1) creatively formulating novel counterfactual operations based on the rich attributes of charts;
(2) professionally composing logically consistent hypothetical questions aligned with the given context.
Empirically, even when provided with detailed task descriptions and examples, LLMs tend to repetitively reference prior data, lacking valuable insights essential for designing diverse HQA data. Therefore, we propose the Counterfactual Proposal Generator to encourage the LLM to generate general instruction proposals that describe common counterfactual operations for different types of charts. Specifically, this component first initializes an instruction proposal pool, in which four seed proposals are manually crafted for each chart type. Then, given a set of general descriptions of chart attributes ${D_C}$, the module utilizes GPT-4 to generate counterfactual instruction proposals $I_P$, formulated as follows:
\begin{equation}
\begin{aligned}
&I_P = \textit{GPT-4}(I_S, P_I, D_C),
\end{aligned}
\end{equation}
where $I_S$ represents a sampled seed proposal from the instruction proposal pool as a contextual example, and $P_I$ is a guiding prompt for GPT-4, with the specific prompt details presented in Appendix~\ref{app:hqa_template}.

After obtaining a diverse set of instruction proposals, the CPG module further utilizes generated proposals to generate specific hypothetical question-answering instances, ensuring their alignment with the context of particular factual chart-based questions. Notably, each hypothetical question corresponds to a factual question from ChartQA, but not every factual question in ChartQA has a corresponding hypothetical question. To generate specific HQA instances, we use the ChartQA annotation JSON file ${A_C}$, which contains structured metadata about the chart’s content. Our empirical findings indicate that such structured textual representations provide LLMs with a more effective understanding of chart information compared to direct image inputs. Subsequently, we design an appropriate prompt to guide GPT-4 in generating specific HQA data based on the previously generated instruction proposals $I_P$, the original factual QA instance $QA_O$, and the chart annotations $A_C$, as follows:
\begin{equation}
QA_H = \textit{GPT-4}(I_P, QA_O, P_H, A_C)
\end{equation}
where $P_H$ represents a prompt template, its specific content detailed in Appendix~\ref{app:hqa_template}.

\subsection{Human-feedback Discriminator}
\label{sec:human}

Although the counterfactual proposal generation component carefully utilizes the LLM to generate HQA data, it is inevitable that some low-quality instances may be produced, containing contradictions with the inherent structure of charts or unreasonable assumptions. To address this issue, we adopt a human validation process to ensure the quality of the generated HQA instances. Specifically, we recruit seven experts with professional knowledge of chart interpretation to review the generated HQA instances. Given the HQA instances produced by the counterfactual hypothesis proposal generation component, the reviewers are required to assess whether the generated counterfactual hypotheses align with the structural properties of the chart and whether the corresponding answers are correct. For instance, as illustrated in Figure~\ref{fig:dataset_generation}, the reviewers evaluate the generated HQA instances from multiple perspectives, including the reasonableness of the question, accuracy of the answer, and complexity of the reasoning process. Based on human judgments, only validated HQA instances are retained, and the corresponding human feedback is incorporated into the instruction proposals before being added back to the instruction proposal pool. As a result, 63.4\% of the generated HQA data successfully passed validation. Specific examples of HQA data review can be found in the appendix~\ref{app:review}.

\begin{table}[t]
\centering
\resizebox{0.45\textwidth}{!}{
\begin{tabular}{l|c|c|c}
\toprule
\multirow{2}{*}{Benchmarks} &
 \multicolumn{1}{c|}{Chart} & \multicolumn{1}{c|}{Question} & \multicolumn{1}{c}{QA formats} \\ 
 \cmidrule(l){2-4}                       
 & Real-world  & Hypothetical  & Open-ended     
 \\ \midrule 
Figure QA  & \textcolor{red}{\XSolidBrush}  & \textcolor{red}{\XSolidBrush}  & \textcolor{red}{\XSolidBrush}   \\
DVQA       & \textcolor{red}{\XSolidBrush}  & \textcolor{red}{\XSolidBrush}  & \textcolor{red}{\XSolidBrush}   \\
LEAF-QA++  & \textcolor{red}{\XSolidBrush}  & \textcolor{red}{\XSolidBrush}  & \textcolor{red}{\XSolidBrush}   \\
PlotQA     & \textcolor{red}{\XSolidBrush}  & \textcolor{red}{\XSolidBrush}  & \textcolor{DarkGreen}{\Checkmark}   \\
ChartLlama & \textcolor{red}{\XSolidBrush}  & \textcolor{red}{\XSolidBrush}  & \textcolor{DarkGreen}{\Checkmark}   \\
MMC    & \textcolor{red}{\XSolidBrush}  & \textcolor{red}{\XSolidBrush}  & \textcolor{DarkGreen}{\Checkmark}   \\
ChartQA    & \textcolor{DarkGreen}{\Checkmark}  & \textcolor{red}{\XSolidBrush}  & \textcolor{DarkGreen}{\Checkmark}   \\
ChartBench    & \textcolor{DarkGreen}{\Checkmark}  & \textcolor{red}{\XSolidBrush}  & \textcolor{DarkGreen}{\Checkmark}   \\
ChartX    & \textcolor{DarkGreen}{\Checkmark}  & \textcolor{red}{\XSolidBrush}  & \textcolor{DarkGreen}{\Checkmark}   \\

\midrule
\textbf{Chart-HQA (ours)} & \textcolor{DarkGreen}{\Checkmark}  & \textcolor{DarkGreen}{\Checkmark}  & \textcolor{DarkGreen}{\Checkmark}   \\
\bottomrule
\end{tabular}
}
\caption{Comparison between existing benchmarks and our new Chart-HQA benchmark.}
\label{comparison}
\vspace{-4mm}
\end{table}
\begin{table}[tp!]
\centering
\resizebox{0.4\textwidth}{!}{
% \scalebox{0.90}{
\begin{tabular}{l r}
\toprule
\textbf{Statistic} & \textbf{Number} \\ 
\midrule
% Total questions & 2388 \\
% \hspace{3mm}* open-ended questions &  2172\\
% \hspace{3mm}* multi-choice questions &  216\\
% \midrule
\# of hypothetical questions & 2173 \\
\# of instruction proposals &  900 \\
\# of charts & 947 \\
% \# of different questions & 23,259 \\
% \# of different answers &  \\
\# of answer types &  4 \\
\midrule
Avg. Character per question &  149.14\\
Avg. Character per assumption &  82.10 \\
Avg. Character per answer &  6.29\\
% Solution length (Average/Max) & 49.5 / 350 \\
 % 4 & w/o Iteration  &\Checkmark& \Checkmark & \Checkmark& \XSolidBrush & 0.1599  & 0.1591\\ 
\bottomrule
\end{tabular}
}
\caption{Key statistics for Chart-HQA.}
\label{tab:statistic}
\vspace{-5mm}
\end{table}
\section{Dataset Analysis}
We apply the proposed data synthesis method to change questions from widely used benchmarks ChartQA~\cite{ChartQA} test-split to be hypothetical by adding a related assumption. The generated HQA benchmark is named Chart-HQA. We present the data analysis for Chart-HQA as below.

\subsection{Comparison to Existing Benchmarks} As shown in Table~\ref{comparison}, Chart-HQA differs from related benchmarks in various aspects: (1) Chart-HQA is the first benchmark to study hypothetical problems over chart context on open domains; (2) Questions in Chart-HQA are generated automatically by LLMs, which greatly reduces data construction costs.
(3) Chart-HQA has an open-vocabulary QA format that requires applying counterfactual operations on the underlying chart data.

\subsection{Key Statistics}
The main statistics for Chart-HQA are shown in Table~\ref{tab:statistic}. The Chart-HQA benchmark contains 2172 hypothetical questions, which are all used to test zero-shot chart-based visual question answering. There are 900 instruction proposals, indicating that Chart-HQA has a rich diversity in the hypothetical problem distribution. The assumptions have an average of 82.1 characters in length, showing that they have lexical richness. There are four answer types in Chart-HQA, and the answer could be a text span, an integer number, a decimal number, or a boolean answer. These statistics suggest that models need diverse symbolic reasoning abilities to answer the questions in Chart-HQA.
\begin{figure}
    \centering
    \small
    \includegraphics[width=0.48\textwidth]{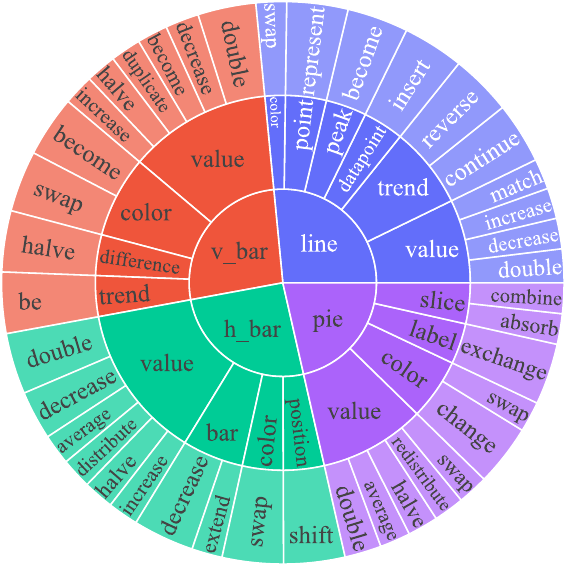} \\
    \caption{Counterfactual operations in generated instruction proposals. The inner circle denotes noun objects in charts, the outer circle represents the action against the noun object.}
    \label{fig:diversity}
    \vspace{-4mm}
\end{figure}

\subsection{Proposal Diversity} 
We further demonstrate the diversity of generated instruction proposals. We identify the counterfactual operations in generated instruction proposals and then extract the verb-noun structure in the counterfactual operation using Berkeley Neural Parser~\cite{kitaev-klein-2018-constituency, kitaev-etal-2019-multilingual}. We randomly parse 10 generated instruction proposals for each chart style. As shown in Figure \ref{fig:diversity}, we can see quite diverse intents and textual formats in these instruction proposals. Notably, the generated counterfactual operations adeptly capture the distinctive characteristics of data visualization. For instance, the counterfactual operation of "reversing trend" for the line chart introduces novel challenges to the model's domain knowledge.

\section{Experiments}

\begin{table*}[thbp!]
\centering

\resizebox{\linewidth}{!}{
\begin{tabular}{l|c|c|cc|cc|c}
\toprule
\multirow{2}{*}{Models} & \multirow{2}{*}{Type} & \multirow{2}{*}{\#Params} & \multicolumn{2}{c|}{ChartQA} & \multicolumn{2}{c|}{Chart-HQA} & \multirow{2}{*}{Decline Rate (\(\downarrow\))} \\ 
\cmidrule(r){4-5} \cmidrule(r){6-7}
 & & & Acc (\(\uparrow\)) & Rank & Acc (\(\uparrow\)) & Rank &  \\ 
\midrule 
\multicolumn{8}{c}{\cellcolor{blue!8}\textbf{Specialist Models}} \\ 
Pix2struct ~\cite{lee2023pix2struct}  & End-to-End & 0.3B  & 56.00  & \#18 & 17.68   & \#17 & 68.43  \\
MatCha ~\cite{liu-etal-2023-matcha}   & End-to-End & 0.3B & 64.20  & \#15  & 21.32   & \#14 & 66.79   \\
Unichart ~\cite{masry-etal-2023-unichart}  & End-to-End & 0.2B & 66.24  & \#13  & 18.69   & \#16  & 71.78  \\
TinyChart~\cite{tinychart}    & End-to-End & 3B  & 83.60  & \#5 &  30.79  & \#11  & 63.17  \\
DocOwl-v2.0~\cite{docowl2}    & End-to-End &  8B & 70.00  & \#10 &  30.83  & \#10  &  55.96 \\
ChartLlama ~\cite{han2023chartllama}   & End-to-End & 13B  & 69.66  & \#11  & 17.22   & \#18 & 75.28  \\
ChartVLM-L~\cite{xia2024chartvlm}    & End-to-End & 14.3B  & 62.28  & \#16 & 20.15   & \#15  & 63.70  \\
DePlot(GPT 3.5) PoT SC~\cite{liu-etal-2023-deplot} & Tool-augment & -  & 76.70  & \#8  & 49.49   & \#6  & 35.48  \\
\midrule
\multicolumn{8}{c}{\cellcolor{blue!8}\textbf{Generalist Models}} \\
Qwen-VL-Chat~\cite{qwen-vl}  & End-to-End & 7B  & 66.30 & \#12  & 28.60   & \#12  & 56.86  \\
DeepSeek-VL-Chat~\cite{lu2024deepseekvl}  & End-to-End & 7B  & 60.72 & \#17  &  27.93  & \#13  & 54.00 \\
Qwen2.5-VL~\cite{Qwen2.5-VL}  & End-to-End & 7B  &  \cellcolor[RGB]{255,250,205}{\textbf{87.30}} & \textbf{\#2}  &  \cellcolor[RGB]{230,246,255}\textbf{57.20}  &  \textbf{\#3} & 34.48  \\
InternVL-v2.5-8B~\cite{internvl}  & End-to-End & 8B  & 84.80 &  \#4 &  48.23  & \#7  &  43.13  \\
Monkey~\cite{li2023monkey}   & End-to-End & 9.8B  & 65.10 & \#14  & 31.45   & \#9  & 51.69  \\
Qwen2.5-VL~\cite{Qwen2.5-VL}  & End-to-End & 72B  & \cellcolor[RGB]{233,252,232}{\textbf{89.50}} & \textbf{\#1 } &  \cellcolor[RGB]{233,252,232}{\textbf{66.41}}  & \textbf{\#1}  & \cellcolor[RGB]{233,252,232}{\textbf{25.80}} \\
% \textbf{InternVL-v2.5-78B}~\cite{internvl}  & End-to-End & 78B  & 88.30 &   &    & 2  &   \\
Gemini-Pro~\cite{team2023gemini}  & End-to-End & -  & 74.10 & \#9  & 41.25   & \#8  & 44.33  \\
Qwen-VL-Max~\cite{qwen-vl-max}  & End-to-End & -  & 79.80 & \#6  & 53.41   & \#5  & 33.07  \\
GPT-4V~\cite{GPT4v}  & End-to-End & -  & 78.50 & \#7  & 56.49   & \#4  & \cellcolor[RGB]{230,246,255}\textbf{28.01}  \\
GPT-4o~\cite{GPT4v}  & End-to-End & -  & \cellcolor[RGB]{230,246,255}\textbf{85.70} & \textbf{\#3}  & \cellcolor[RGB]{255,250,205}\textbf{62.52}   & \textbf{\#2}  & \cellcolor[RGB]{255,250,205}\textbf{27.05} \\
% \textbf{{Ours} (GPT-4V)} & &&\textbf{80.40} && \textbf{60.59}&& \\
\bottomrule 
\end{tabular}}
\caption{Zero-shot transfer results with state-of-the-art generalist multi-modal language methods and chart-oriented specialist models on our proposed Chart-HQA. PoT denotes program-of-thought prompting. SC denotes self-consistency. We color each column as the \ctext[RGB]{233,252,232}{best}, \ctext[RGB]{255,250,205}{second best}, and \ctext[RGB]{230,246,255}{third best}.}
\label{tab:hqa}
\vspace{-3mm}
\end{table*}

\label{sec:experiment}
In this section, we conduct zero-shot transfer experiments on the proposed Chart-HQA (Section~\ref{sec:zero}).  
In addition, we further perform a fine-grained analysis of MLLMs based on answer types (Section~\ref{sec:answer}).  Before discussing results, we provide details of the experimental setup below.

\subsection{Experimental Settings}
\label{sec:setup}
\noindent\textbf{Models.}
% We evaluate two types of models: (1) chart-oriented specialist models. This type of model specializes in pre-training on large amounts of chart data. we evaluate end-to-end models including Pix2struct~\cite{lee2023pix2struct}, MatCha ~\cite{liu-etal-2023-matcha}, Unichart~\cite{masry-etal-2023-unichart}, ChartLlama~\cite{han2023chartllama} and ChartVLM~\cite{xia2024chartvlm}. We also evaluate a tool-augmented model Deplot~\cite{liu-etal-2023-deplot}. 
% We use the GPT3.5~\cite{openai2022chatgpt} as the inference model of Deplot for our experiments. 
% (2) generalist models, which are trained towards general capability for various vision-language tasks. The open-source models include Monkey~\cite{li2023monkey}, Qwen-VL-Chat~\cite{qwen-vl}, while the closed-source models contain Qwen-VL-Max~\cite{qwen-vl-max}, Gemini-Pro \cite{team2023gemini}, and GPT-4V \cite{GPT4v}. 
% For open-source models, we re-implement the results using the official codes. For closed-source models, we re-implement the results using the official APIs. 
We evaluate two types of models: (1) chart-oriented specialist models. This type of model specializes in pre-training on large amounts of chart data. we evaluate end-to-end models including Pix2struct, MatCha , Unichart, ChartLlama, ChartVLM, TinyChart and Docowl2.0. We also evaluate a tool-augmented model Deplot
\footnote{We use the GPT3.5~\cite{openai2022chatgpt} as the inference model of Deplot for our experiments.}. (2) generalist models, which are trained towards general capability for various vision-language tasks. The open-source models include Monkey, Qwen-VL-Chat, DeepSeek-VL, Qwen2.5-VL and InternVL-2.5, while the closed-source models contain Qwen-VL-Max, Gemini-Pro, GPT-4V, GPT-4o. 
For open-source models, we re-implement the results using the official codes. For closed-source models, we re-implement the results using the official APIs.  

\noindent\textbf{Metrics.} 
To ensure a fair comparison with ChartQA results, we adopt the same evaluation method and metric used in ChartQA. Specifically, we choose the \textit{relaxed accuracy} used in ChartQA as the evaluation metric, which means exact match accuracy with 5\% tolerance on numerical error is used to report all QA results. In addition, we compute the \textit{decline rate} to measure the performance difference of models between ChartQA and Chart-HQA, which is calculated as follows:  
\begin{equation}
\text{Decline Rate} = \frac{\left| Acc_{QA} - Acc_{HQA} \right|}{Acc_{QA}} \times 100\%.
\end{equation}

% We evaluate the proposed approach on four chart-based visual question answering tasks, including factoid QA, fact-checking, chart summarization, and proposed hypothetical QA. (1) For factoid QA, we evaluate the average performance 
% using relaxed accuracy (exact match but tolerating 5\% of numerical
% error) on ChartQA~\cite{chartqa}.
% (2) For fact checking, we evaluate macro F1 score on ChartCheck~\cite{akhtar2024chartcheck} benchmark.
% (3) For the summarization task, we use the Chart-to-text~\cite{chart-to-text} dataset. We follow the evaluation framework and scoring criteria of ChartLlama~\cite{han2023chartllama} and report the GPTscore. 
% (4) For hypothetical question answering, we evaluate zero-shot reasoning performance on the proposed Chart-HQA. We choose the relaxed accuracy used in ChartQA as the evaluation metric.

\subsection{Zero-shot Transfer on Chart-HQA}
\label{sec:zero}
\textbf{Chart-HQA establishes a highly challenging benchmark for visual chart understanding.} Table~\ref{tab:hqa} compares various MLLMs on the ChartQA and our Chart-HQA.
First, chart-specialist models exhibit severe generalization issues when handling counterfactual assumptions added to questions while keeping chart images unchanged.
For example, large-scale specialist models with over 10B parameters, such as ChartVLM-L and ChartLlama, exhibit significant declines of 63.70\% and 75.28\% respectively on Chart-HQA compared to their strong performance on ChartQA. Second, generalist MLLMs also demonstrate limited capabilities in counterfactual reasoning over charts.  
For instance, the high-performing GPT-4o exhibits a notable 27.05\% decline rate on Chart-HQA compared to its performance on ChartQA. 
% This further underscores the challenge of answering hypothetical questions in visual chart reasoning tasks. 
Third, we further find that enhancing the model’s symbolic reasoning ability is crucial for Chart-HQA. For example, with the same model size ($\sim$7B), Qwen2.5-VL significantly outperforms Qwen-VL-Chat and InternVL-2.5 on Chart-HQA.

\begin{table*}[thbp]
    \centering
    \resizebox{\textwidth}{!}{
    \begin{tabular}{l|c|c|cccc|c}
    \toprule
    \multicolumn{1}{c|}{\multirow{2}{*}{Model}} & \multirow{2}{*}{Type} & \multirow{2}{*}{\#Params} & \multicolumn{4}{c|}{Chart-HQA} & \multirow{2}{*}{Variance (\textbf{$\downarrow$})} \\ 
    \cmidrule(l){4-7}                       
     & & & INT  & DEC & BOOL & TEXT &  \\ 
    \midrule 
    \multicolumn{8}{c}{\cellcolor{blue!8}\textbf{Specialist Models}} \\ 
    Pix2struct~\cite{lee2023pix2struct}      & End-to-End & 0.3B  & 17.81 & 18.48 & 20.21 & 14.60  & \cellcolor[RGB]{255,250,205}\textbf{4.13} \\
    MatCha ~\cite{liu-etal-2023-matcha}      & End-to-End & 0.3B  & 23.01 & 21.83 & 4.26  & 20.94  & 59.06 \\
    Unichart  ~\cite{masry-etal-2023-unichart} & End-to-End & 0.2B  & 15.21 & 15.74 & 40.43 & 28.10  & 107.30 \\
    TinyChart  ~\cite{tinychart} & End-to-End & 3B  & 35.71 & 26.61 & 57.45 & 39.39  & 125.60 \\
    DocOwl-v2.0  ~\cite{docowl2} & End-to-End & 8B  & 30.36 & 27.84 & 54.26 & 37.47  & 106.30 \\
    ChartLlama ~\cite{han2023chartllama}      & End-to-End & 13B   & 13.27 & 11.88 & 59.57 & 28.73 & 368.37 \\
    ChartVLM-L~\cite{xia2024chartvlm}        & End-to-End & 14.3B & 17.99 & 16.62 & 51.09 & 26.26  & 191.47 \\
    DePlot(GPT 3.5) PoT SC~\cite{liu-etal-2023-deplot} & Tool-augment & -  & 45.08 & 58.17 & 56.38 & 32.96  & 102.07 \\
    \midrule
    \multicolumn{8}{c}{\cellcolor{blue!8}\textbf{Generalist Models}} \\ 
    Qwen-VL-Chat~\cite{qwen-vl}              & End-to-End & 7B    & 27.53 & 25.71 & 44.68 & 34.44 & 55.38 \\
    DeepSeek-VL-Chat~\cite{lu2024deepseekvl}  & End-to-End & 7B  & 29.17 & 24.55  & 42.55   & 38.02  & 50.29 \\
Qwen2.5-VL~\cite{Qwen2.5-VL}  & End-to-End & 7B  &  44.05 & 58.66  & \cellcolor[RGB]{255,250,205}\textbf{67.02}   & 54.55  & 68.35  \\
InternVL-v2.5-8B~\cite{internvl}  & End-to-End & 8B  & 36.90 & 48.71  &  \cellcolor[RGB]{230,246,255}\textbf{60.64}  & 48.21  &  70.50  \\
    Monkey~\cite{li2023monkey}               & End-to-End & 9.8B  & 30.14 & 28.32 & 44.68 & 39.12 & 44.41 \\
Qwen2.5-VL~\cite{Qwen2.5-VL}  & End-to-End & 72B  & \cellcolor[RGB]{255,250,205}\textbf{51.55} & \cellcolor[RGB]{233,252,232}\textbf{68.02}  & \cellcolor[RGB]{233,252,232}\textbf{72.34}   &  \cellcolor[RGB]{233,252,232}\textbf{64.54} & 60.24  \\
% \textbf{InternVL-v2.5-78B}~\cite{internvl}  & End-to-End & 78B  & 88.30 &   &    & 2  &   \\
    Gemini-Pro~\cite{team2023gemini}         & End-to-End & -     & 38.71 & 43.55 & 55.32 & 36.46 & 53.06 \\
    Qwen-VL-Max~\cite{qwen-vl-max}           & End-to-End & -     & \cellcolor[RGB]{230,246,255}\textbf{50.34} & 55.98 & 58.51 & 49.72 & \cellcolor[RGB]{230,246,255}\textbf{13.86} \\
    GPT-4V~\cite{GPT4v}                      & End-to-End & -     & \cellcolor[RGB]{233,252,232}\textbf{54.58} & \cellcolor[RGB]{230,246,255}\textbf{58.38} & 55.32 & \cellcolor[RGB]{230,246,255}\textbf{\textbf{55.52}} & \cellcolor[RGB]{233,252,232}\textbf{2.09} \\
    GPT-4o~\cite{GPT4v}  & End-to-End & -  & 49.10 & \cellcolor[RGB]{255,250,205}\textbf{64.79}  &  51.06  & \cellcolor[RGB]{255,250,205}\textbf{61.98}  & 45.72  \\
    \bottomrule
    \end{tabular}}
    \caption{Fine-grained Evaluation Results on Chart-HQA across different answer types. INT: Integer answers; DEC: Decimal answers; BOOL: Boolean text answers; TEXT: Text answers. We color each column as the \ctext[RGB]{233,252,232}{best}, \ctext[RGB]{255,250,205}{second best}, and \ctext[RGB]{230,246,255}{third best}.}
    \label{tab:answer}
    \vspace{-2mm}
\end{table*}
\begin{figure*}[ht]
    \centering
    % \small
    \includegraphics[width=\linewidth]{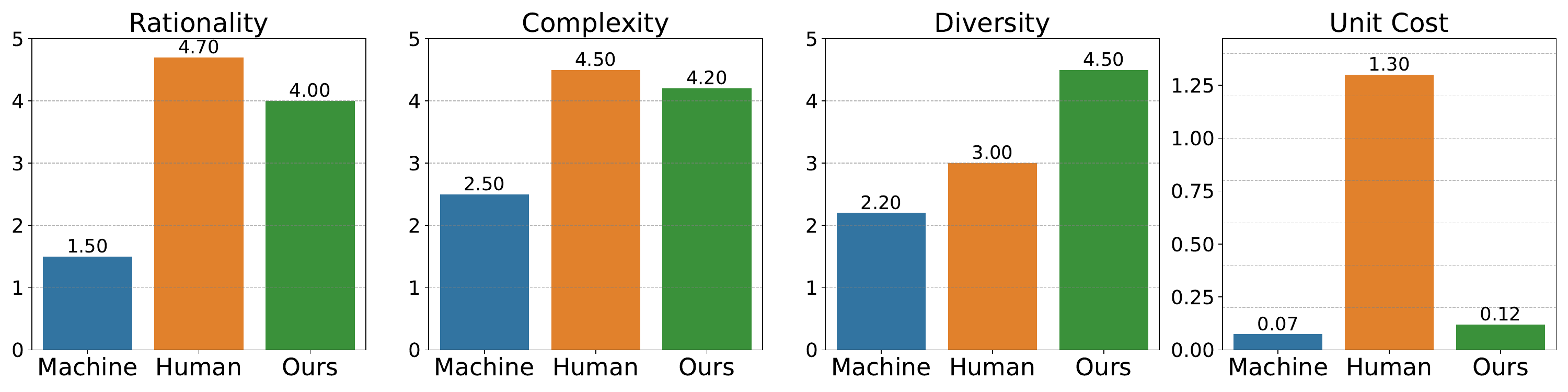}
    \caption{Human evaluation performance of three data synthesis methods, including human, machine, and our human-machine interaction approach. From left to right, the comparison includes question rationality, complexity, diversity, and synthesis cost (unit: CNY).}
    \label{hqa:ablation}
    \vspace{-4mm}
\end{figure*}
\begin{figure*}[t]
    \centering
    % \small
    \includegraphics[width=\linewidth]{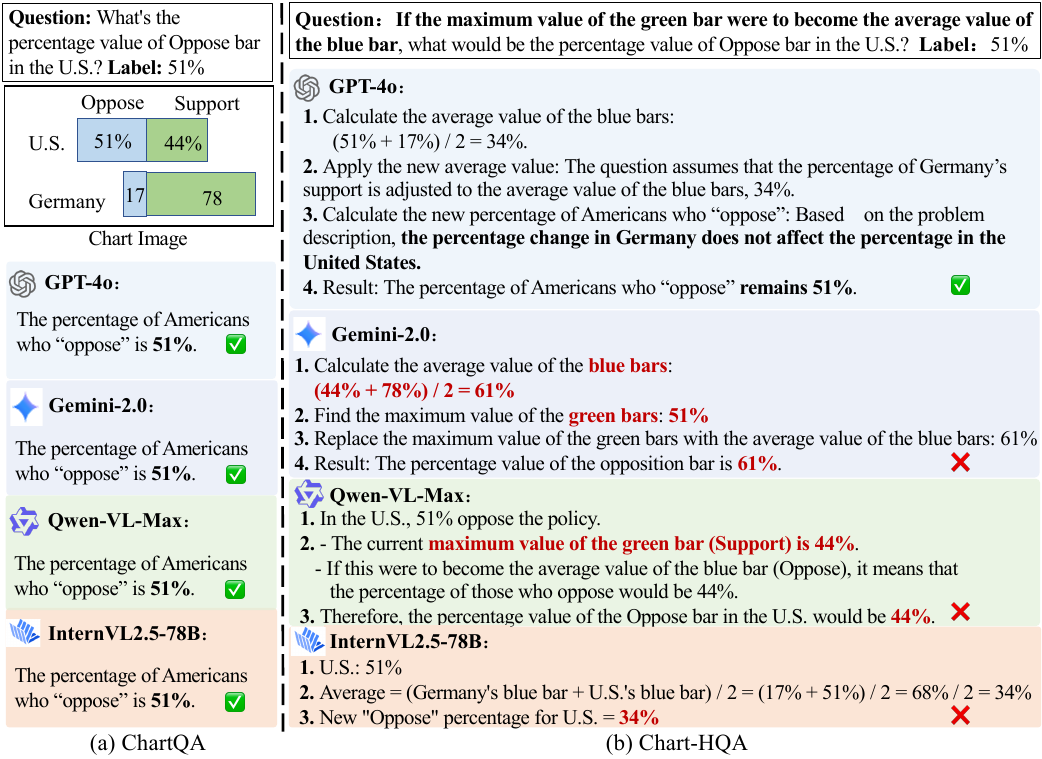}
    \caption{The visualization of examples in ChartQA and Chart-HQA(ours). We use \textbf{black bold} to highlight key reasoning steps of the model and \textcolor{red}{red} to mark incorrect reasoning steps.}
    \label{hqa:case}
    \vspace{-4mm}
\end{figure*}
\subsection{Fine-grained Evaluation Results}
\label{sec:answer}
\textbf{Most MLLMs exhibit imbalanced performance across different answer types within Chart-HQA.} Table~\ref{tab:answer} presents the fine-grained evaluation results on Chart-HQA for different answer types.
First, only GPT-4V demonstrates both high HQA performance across various answer types and balanced performance distribution. For example, GPT-4V achieves the best performance among all evaluated models on integer, decimal, and boolean answer types, and its performance variance (2.09) is the lowest among all evaluated models. Second, generalist models exhibit superior fine-grained performance and reasoning stability compared to chart-specialist models. This phenomenon indicates that pretraining on fundamental general-purpose abilities is beneficial for chart understanding.

% This further suggests that pretraining on fundamental general-purpose abilities remains beneficial for vertical-domain tasks such as chart-based question answering. 
% This indicates that GPT-4V possesses comprehensive symbolic reasoning capabilities, including logical reasoning, mathematical reasoning, and natural language reasoning.  

\subsection{Ablation Study}
\vspace{-1.5mm}
We further investigate the effectiveness of our HQA data synthesis method.   
Specifically, we synthesized 100 HQA instances respectively using three different approaches:  
Human expert-designed (Human), LLM-generated (Machine) and our proposed human-AI interactive method.
Subsequently, we calculate the unit cost of data synthesis for each method.  
We then invite human experts to evaluate the synthesized data from three perspectives, each scored on a scale of 5:  \textit{Rationality}, which measures whether the generated questions conform to the intrinsic layout structure of the chart; 
\textit{Complexity}, which assesses the difficulty of answering the questions; and  \textit{Diversity}, which evaluates the richness of counterfactual operations applied to the chart.  
The evaluation results are presented in Figure~\ref{hqa:ablation}.  
First, in terms of rationality and complexity,  our method performed comparably to human experts. For example, in complexity scoring, human-designed questions received a score of 4.5, while our method achieved a close score of 4.2. Second, regarding diversity, our method significantly outperform both human-designed and machine-generated approaches. Third, compared to the design cost of human experts, our method reduces costs by 90.7\%, lowering the average unit cost to 0.12 CNY per sample. The primary reason for this cost efficiency is that, compared to the direct design HQA instances by human experts, our method effectively distributes the workload, where the LLM generate the questions and human experts review them,  
thereby significantly reducing the overall cost. 

\subsection{Case Study}
To better illustrate the challenge of the proposed chart-based hypothetical question answering task, we conduct a specific case study on Chart-HQA using the powerful reasoning models GPT-4o, Gemini-2.0~\cite{team2023gemini}, Qwen-VL-Max, and InternVL2.5-78B~\cite{internvl} as shown in Figure~\ref{hqa:case}. All models correctly answer the original factual question shown in Figure~\ref{hqa:case}(a).
However, when answering the hypothetical question proposed in Chart-HQA,  
the reasoning processes of models exhibit significant differences as shown in Figure~\ref{hqa:case}(b). Specifically, GPT-4o first accurately reasons through the hypothetical scenario,  
correctly inferring that the maximum value of the green bar (78\%) becomes the average value of the blue bar (34\%).  
It then deduces that the hypothetical scenario does not affect the answer,  
thus maintaining the original response (51\%). In contrast, Gemini-2.0 makes two critical errors. First, it confuses the colors in the chart, leading to an incorrect interpretation of the hypothetical scenario. 
Second, it fails to reason that the hypothetical scenario does not influence the question, ultimately replacing the correct answer with the erroneous assumption inference. Similar to Gemini-2.0, Qwen-VL-Max and InternVL2.5-78B also fail to recognize that the hypothetical scenario does not affect the answer during the reasoning process, leading to incorrect responses. This phenomenon indicates that since MLLMs need to analyze the problem step by step, they tend to become deeply engaged in imaginative reasoning based on the assumed scenario, ultimately overlooking the actual content the question aims to query.

\section{Related Works}
\subsection{Chart Benchmarks}
HallusionBench~\cite{HallusionBench} has revealed that state-of-the-art models, such as GPT-4V~\citep{GPT4} and LLaVA-1.5~\citep{Improvedllava}, exhibit severe hallucinations when processing intricate chart-related queries. Additionally, several benchmarks including SciCap~\citep{SciCap}, Chart2Text~\cite{Chart2Text}, AutoChart~\cite{AutoChart}, and ChartSumm~\cite{ChartSumm}, focus on chart-to-text summarization. For chart comprehension, ChartQA~\cite{ChartQA} and PlotQA~\cite{PlotQA} serve as widely used evaluation datasets. ChartLlama~\cite{Chartllama}, ChartX~\cite{ChartX}, ChartY~\cite{OneChart} and ChartBench~\cite{chartbench} significantly increases the number of supported chart types and dataset scale.
In contrast, we introduce Chart-HQA to systematically analyze the impact of inherent output biases in MLLMs on chart-based evaluations. By comparing its performance with the widely used ChartQA, we reveal the limitations of current MLLMs in visual chart understanding.

\section{Conclusion}
In this paper, we propose an novel chart hypothetical question answering task to reveal the inherent output bias problem of MLLMs. Subsequently, we present a human-machine interactive HQA data synthesis framework named {\model} to synthesize diverse and high-quality HQA data at a low cost. We synthesized a challenging benchmark called Chart-HQA using publicly available data sources. Through a comprehensive analysis of 18 MLLMs of varying sizes, we reveal the shortcomings of current MLLMs in visual chart understanding.

% an automatic chain-of-tool prompts optimization framework named {\model} to reliably optimize chain-of-tool prompts to complete various chart-based visual question answering tasks. Furthermore, we extend traditional factoid QA to hypothetical question answering (HQA). We develop a general multi-stage approach to automatically generate an HQA dataset, named Chart-HQA, to promote this challenging direction. Extensive experiments validate that {\model} not only outperforms existing methods in traditional chart-based visual question answering tasks but also exhibits superior transfer performance on the more challenging Chart-HQA benchmark. 
\section*{Limitations}
% The proposed work still contains several limitations to address in future work, as follows:\\
% \noindent \textbf{Method.}
% The limitations of our method are that it cannot invoke external knowledge bases, such as obtaining relevant knowledge through an Internet browser. Due to the frequent occurrence of proprietary terms in chart problems, retrieving external knowledge is necessary.
% We leave this for future work.

% % One limitation of our method is that it cannot capture global structure information from informative visual clues. The visual features in visually-rich documents such as font size and color may provide diverse structure information. For example, section titles in resumes and job ads are often in fonts different from the content. We leave this for future work.
% \\
% \noindent \textbf{Benchmark.}
% %  Due to the limited budget and computation resources，我们只在Chart-hqa上进行了零样本测试。未来我们会增大数据规模探索更多的实验设置
Due to the limited budget and computation resources, we only conducted zero-shot testing on Chart-HQA. In the future, we will increase the data scale to explore more experimental settings.
% We only evaluate the visual relation extraction task covering diverse settings. Due to the limited budget and computation resources, we cannot afford evaluation on more tasks related to visually-rich documents. We will plan to evaluate the proposed approach on more visual information extraction tasks such as semantic entity recognition.

\section*{Ethical Statement}
The dataset in this paper is constructed using publicly available sources and adheres to ethical guidelines for data collection and annotation. No personally identifiable or sensitive information is included. Efforts have been made to ensure fairness and minimize bias in data representation.
\bibliography{citation}
\bibliographystyle{acl_natbib}
\clearpage
\appendix
% \section{Extended Related Work}
% \label{app:related_work}
% \subsection{Multi-Modal Chart Reasoning}
% Traditionally, chart-based visual question answering tasks are dominated by end-to-end neural models with chart-specific architectural designs~\cite{liu-etal-2023-matcha,lee2023pix2struct,masry-etal-2023-unichart,han2023chartllama}. Recently, there has been a surge in 
% calling external tools to process charts, and leveraging the powerful symbolic reasoning capability of large language models for downstream tasks such as QA. DePlot~\cite{liu-etal-2023-deplot} first converts the chart to a table, after which it prompts the LLM to reason over the translated text. DOMINO~\cite{domino} devised a dual-system, which alternates between system-2 (a prompted LLM) and system-1 (a visual encoder-text decoder) to tackle intricate questions regarding charts. Different from these approaches, {\model} proposes a multi-agent system architecture and automatically optimizes multi-agent prompts adapted to different chart reasoning tasks.

% \subsection{Initial agent prompts of {\model}}
% \label{app:mutli-agent}
% In this section, we show the initial agent prompts for ChartQA, ChartCheck and Chart-to-Text respectively in Table \ref{agent:prompt_qa}, ~\ref{agent:prompt_check}, ~\ref{agent:prompt_text}.  

\section{Prompts Templates for HQA}
\label{app:hqa_template}
Within this section, we outline the prompt templates for automatically generating hypothetical questions. The prompt templates are shown in Table~\ref{tab:prompt_ins_gen}, ~\ref{tab:prompt_hq_gen}.
\begin{table*}[h]\centering
\begin{minipage}{0.9\textwidth}
%\vspace{0mm}    
\centering
\caption{Prompt template of instruction proposal synthesis.}
\begin{tcolorbox} 
    \centering
     %\hspace{-4mm}
      \small
    \begin{tabular}{p{1.0\textwidth}}
   \textcolor[rgb]{0.8,0.3,0}{ {\bf SYSTEM:} } \\ 
   You are a creative prompt creator.  \\
  \textcolor[rgb]{0.8,0.3,0}{ {\bf USER:} } \\ 
    Given \{CHART\_DESCRIPTION\}.\\
    A series of data points contains a list of the following attributes (dictionary-style): \\
    \{FIELD\_DESCRIPTION\} \\
    According to the chart description provided above, Your goal is to generate new instructions to guide the user in asking hypothetical questions based on information in the chart.\\
    Your can draw inspiration from the \#Given Instructions\# to create a brand new instruction. \\
    The new instruction must meet the following conditions: \\
    1. It should only contains two parts: how to specify the elements and the assumed change to be applied on the elements.\\
    2. The new instruction must be reasonable and must be understood and responded by humans. \\
    3. Follow the sentence patterns in the examples.\\
    4. Please replace specific concepts with general concepts.\\
    5. Use attributes in charts to refer to specific elements.\\
    \#Given Instructions\#:\\
    1. \{I1\}\\
    2. \{I2\}\\
    3. \{I3\}\\
    4. \{I4\}\\
    Now please directly generate 3 new instructions without writing any other explanations:\\
   \textcolor[rgb]{0.8,0.3,0}{ {\bf <Output>:}  } 
    \end{tabular}
    \label{hqa:proposal}
\end{tcolorbox}
    \label{tab:prompt_ins_gen}
\end{minipage}
\end{table*}
\section{Illustrative Examples of Human Verification for HQA}
\label{app:review}
\begin{figure*}[h]
    \centering
    \small
    \includegraphics[width=0.6\textwidth]{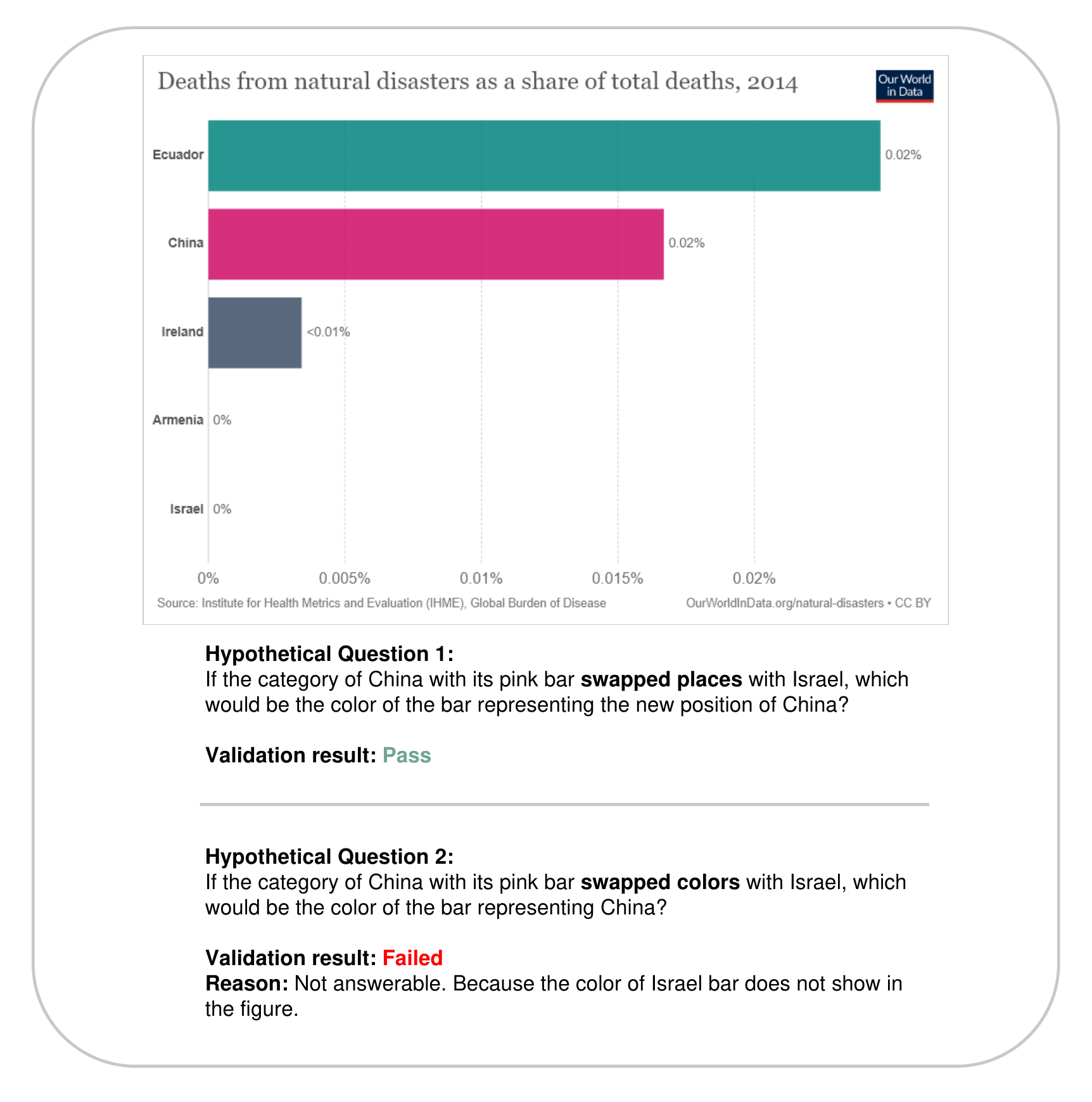} \\
    \caption{The first example of human verification.}
    \label{fig:human_verification_case1}
\end{figure*}

\begin{figure*}[h]
    \centering
    \small
    \includegraphics[width=0.6\textwidth]{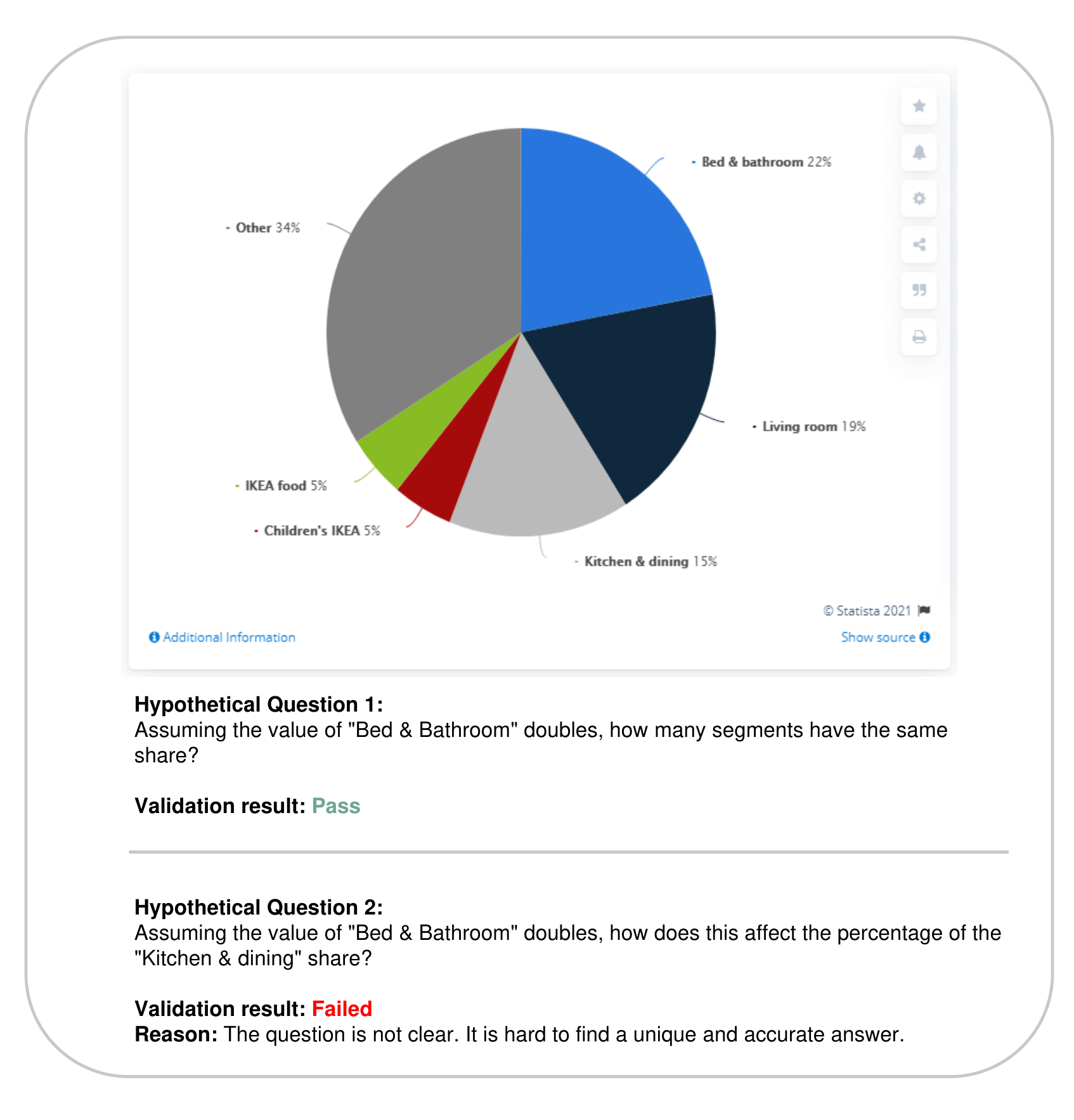} \\
    \caption{The second example of human verification.}
    \label{fig:human_verification_case2}
\end{figure*}

Within this section, we show the examples of human verification processes for ensuring the quality of generated hypothetical questions. These examples are illustrated in Figures ~\ref{fig:human_verification_case1}, ~\ref{fig:human_verification_case2}.
% \section{}
% \label{app:case}
% Within this section, we show the examples of human verification processes for ensuring the quality of generated hypothetical questions. These examples are illustrated in Figures ~\ref{fig:human_verification_case1}, ~\ref{fig:human_verification_case2}.

% \clearpage

\begin{table*}[th!]\centering
\begin{minipage}{0.8\textwidth}
%\vspace{0mm}    
\centering
\caption{Prompt template for hypothetical question generation.}
\begin{tcolorbox} 
    \centering
     %\hspace{-4mm}
      \small
    \begin{tabular}{p{0.95\textwidth}}
   \textcolor[rgb]{0.8,0.3,0}{ {\bf SYSTEM:} } \\ 
   You are a Question Rewriter.  \\ 
  \textcolor[rgb]{0.8,0.3,0}{ {\bf USER:} } \\ 
    You are provided with metadata from \{CHART\_DESCRIPTION\}. The chart's title and series of data points (models) are given in the metadata, with each model comprising attributes outlined in \{FIELD\_DESCRIPTION\}.\\
    Your role is to creatively rewrite original questions into Hypothetical Questions (HQ) based on the chart's information. Each original question should be rephrased into two different hypothetical questions.\\
    Ensure:\\
    1. Adhere to the ideas in \#Feasible Rewrite Proposals\#.\\
    2. HQ should also adhere to the format in \#Demonstration\# and use specific details from the chart. It also needs to be as clear as possible.\\
    3. Keep the original question as part of rewritten HQ.\\
    4. The answer to the HQ should differ from the original answer.\\
    5. Provide the name of the color in words, not any code like \#FF0000.\\
    6. When the answer is a percentage value, it needs to be answered as a percentage.\\
    7. If the calculation process includes percentage values, you need to pay attention to the percent operation.\\
    \\
    \#Feasible Rewrite Proposals\#\\
    1. \{I1\}\\
    2. \{I2\}\\
    3. \{I3\}\\
    \\
    \#Demonstration\#:\\
    Original question: \{Q\_DEMON\}\\
    Hypothetical question examples:\\
    1. \{HQ\_DEMON\_1\}\\
    2. \{HQ\_DEMON\_2\}\\
    \\
    \#Chart Metadata\#:\\
    \{CHART\_METADATA\}\\
    \\
    **Please directly complete HQs and produce the following text information. Note that the answers should not include any explanation or units.**:\\
    First Original Question:\\
    Question: \{Q1\}\\
    Answer: \{A1\}\\
    HQ Rewrites:\\
    Question\_1: \\
    Answer\_1: \\
    Question\_2: \\
    Answer\_2: \\
    Second Original Question:\\
    Question: \{Q2\}\\
    Answer: \{A2\}\\
    HQ Rewrites:\\
    Question\_1: \\
    Answer\_1: \\
    Question\_2: \\
    Answer\_2: \\
   \textcolor[rgb]{0.8,0.3,0}{ {\bf <Output>:}  } \\ 
    \end{tabular}
    \label{hqa:generation}
\end{tcolorbox}
%\vspace{-2mm}
    \label{tab:prompt_hq_gen}
\end{minipage}
\end{table*}

\clearpage

\end{document}